\documentclass[acmsmall]{acmart}
\usepackage{multirow}

\AtBeginDocument{%
  \providecommand\BibTeX{{%
    \normalfont B\kern-0.5em{\scshape i\kern-0.25em b}\kern-0.8em\TeX}}}

\copyrightyear{2018}
\acmYear{2018}
\setcopyright{acmlicensed}
\acmConference[Woodstock '18]{Woodstock '18: ACM Symposium on Neural Gaze Detection}{June 03--05, 2018}{Woodstock, NY}
\acmBooktitle{Woodstock '18: ACM Symposium on Neural Gaze Detection, June 03--05, 2018, Woodstock, NY}
\acmPrice{15.00}
\acmDOI{10.1145/1122445.1122456}
\acmISBN{978-1-4503-9999-9/18/06}

\begin{document}

%
% The "title" command has an optional parameter, allowing the author to define a "short title" to be used in page headers.
\title{Instance-Based Classification through  Hypothesis Testing}

%
% The "author" command and its associated commands are used to define the authors and their affiliations.
% Of note is the shared affiliation of the first two authors, and the "authornote" and "authornotemark" commands
% used to denote shared contribution to the research.
\author{Zengyou He}
\email{zyhe@dlut.edu.cn}
\author{Chaohua Sheng}
\author{Yan Liu}
\affiliation{%
  \institution{School of Software, Dalian University of Technology}
  \city{Dalian, China}
}

\author{Quan Zou}
\email{zouquan@nclab.net}
\affiliation{%
  \institution{Institute of Fundamental and Frontier Sciences, University of Electronic Science and Technology of China}
  \city{Chengdu, China}
}

%
% By default, the full list of authors will be used in the page headers. Often, this list is too long, and will overlap
% other information printed in the page headers. This command allows the author to define a more concise list
% of authors' names for this purpose.
\renewcommand{\shortauthors}{He and Sheng, et al.}

%
% The abstract is a short summary of the work to be presented in the article.
\begin{abstract}
Classification is a fundamental problem in machine learning and data mining. During the past decades, numerous classification methods have been presented based on different principles. However, most existing classifiers cast the classification problem as an optimization problem and do not address the issue of statistical significance. In this paper, we formulate the binary classification problem as a two-sample testing problem. More precisely, our classification model is a generic framework that is composed of two steps. In the first step, the distance between the test instance and each training instance is calculated to derive two distance sets. In the second step, the two-sample test is performed under the null hypothesis that the two sets of distances are drawn from the same cumulative distribution. After these two steps, we have two \emph{p}-values for each test instance and the test instance is assigned to the class associated with the smaller \emph{p}-value. Essentially, the presented classification method can be regarded as an instance-based classifier based on hypothesis testing. The experimental results on 40 real data sets show that our method is able to achieve the same level performance as the state-of-the-art classifiers and has significantly better performance than existing testing-based classifiers. Furthermore, we can handle outlying instances and control the false discovery rate of test instances assigned to each class under the same framework.
\end{abstract}

\begin{CCSXML}
<ccs2012>
<concept>
<concept_id>10010147.10010257.10010293.10010315</concept_id>
<concept_desc>Computing methodologies~Instance-based learning</concept_desc>
<concept_significance>500</concept_significance>
</concept>
</ccs2012>
\end{CCSXML}

\ccsdesc[500]{Computing methodologies~Instance-based learning}

%
% Keywords. The author(s) should pick words that accurately describe the work being
% presented. Separate the keywords with commas.
\keywords{classification,hypothesis testing, two-sample testing, machine learning}

%
% A "teaser" image appears between the author and affiliation information and the body 
% of the document, and typically spans the page. 
%%  \includegraphics[width=\textwidth]{sampleteaser}
%%  \caption{Seattle Mariners at Spring Training, 2010.}
%%  \Description{Enjoying the baseball game from the third-base seats. Ichiro Suzuki preparing to bat.}

%
% This command processes the author and affiliation and title information and builds
% the first part of the formatted document.
\maketitle

\section{Introduction}
Classification is a fundamental data analysis procedure, which is ubiquitously used across different fields. Thousands of classification algorithms (classifiers) have been developed during the past decades \citep{JMLRdelgado14a}. These classifiers range from simple models such as \emph{k}-nearest neighbors (\emph{k}-NN) \cite{TITknn} to more sophisticated models such as support vector machine (SVM) \citep{cortes95supportvectornetworks} and random forests (RF) \citep{breiman2001random}.

Despite the advances on the development of new classifiers, no single classification algorithm can always achieve the best performance on all data sets \citep{JMLRdelgado14a}. This indicates that different classifiers are complementary to each other in different contexts. Therefore, it is still necessary to develop new and alternative classifiers based on some principles that remain unexplored.

The motivation behind this research is based on the following observations. First, existing non-lazy classifiers typically formulate the classification problem as an optimization problem. Such optimization-based learning strategies can always generate the target classifiers, regardless of the statistical significance of learnt models. Second, classifiers such as logistic regression are able to provide probability values for categorizing an unknown test instance. However, it is not an easy task to determine a universal probability threshold to ensure that the classification of the test instance into the corresponding class is statistically significant. Last but not least, existing classifiers cannot control the number of misclassified test instances in terms of metrics such as the false discovery rate (FDR). Such capability is quite important in the scenario of biological data analysis, in which the prediction results will be further validated by wet-lab experiments that can be costly and time-consuming \citep{Wagih2015MIMP}. Thus, we need to add some notion of statistical significance to classifiers. 

In fact, the classification problem has already been formulated as a hypothesis testing issue in \citep{liao2007test}. More recently, several research efforts \citep{ghimire2012classification,guo2018interpoint} further extend the initial formulation in \citep{liao2007test} from different aspects. However, the following observations motivate this research. First of all, existing testing-based classification methods deserve certain theoretical drawbacks, as discussed and summarized in Section \ref{Related work}. Second, only simulation data sets and several small real data sets have been empirically tested, making it difficult to convince people on the practical usage of such testing-based formulation. Third, the connection between this new formulation and existing classification methods have never been discussed. Finally, the potential benefit of the testing-based classification model remains unexplored. 

Based on the above observations, we present a new testing-based classification formulation, in which the null hypothesis is that, informally, the test instance does not belong to any class. To precisely define the null hypothesis, we focus on the classification problem in a two-class setting. First, we can calculate the distance between the test instance and each training instance in the training data set. In this way, we will generate two sets of distances for one test instance that needs to be classified. Then, the hypothesis testing issue can be casted as a two-sample testing problem \citep{gibbons2011nonparametric}, in which each sample corresponds to a set of distances. In this formulation, the null hypothesis is that two sets of distances are drawn from the same cumulative distribution.

Two-sample testing is a fundamental problem in statistics. We employ the classical Wilcoxon-Mann-Whitney (WMW) test for quantifying the statistical significance in terms of \emph{p}-values. To alleviate the effect of outlying and irrelevant training instances, we further apply the WMW test to two distance sets that are generated from \emph{k}-NNs of the test instance.

The testing-based classification formulation has several salient features. First of all, it can provide \emph{p}-values for each test instance to quantify the statistical significance of classifying this instance to certain classes. Accordingly, we can detect outlying test instances that do not belong to any class if the \emph{p}-values with respect to all classes are larger than the significance level threshold. Second, we can control the FDR of test instances that are assigned to each class based on their \emph{p}-values. 

We evaluate our method on forty data sets from the UCI \citep{Dua2017} repository and the KEEL-dataset repository \citep{Alcal2011KEEL} with respect to the standard classification task. The experimental results show that our method is able to achieve the same level performance as the state-of-the-art classifiers. Meanwhile, it can handle outlying test instances and control the FDR of test instances assigned to each class in a natural manner. 

The main contributions of this paper can be summarized as follows. 

(1) The binary classification issue is formulated as a two-sample testing problem. Since two-sample testing is a fundamental problem in statistics and many well-known tests are available in the literature, it can be expected that we may introduce many effective testing-based classifiers in the near future. 

(2) The classification model that integrates hypothesis testing and the \emph{k}-NN method is presented. This formulation can alleviate the effect of outlying and irrelevant training instances to improve the classification accuracy significantly. 
 
(3) A comprehensive performance comparison over 40 real data sets is conducted. The experimental results demonstrate the fact that the testing-based classifier is able to achieve the same level performance as standard classifiers such as SVM and decision tree. 

(4) Some interesting connections between our testing-based classifiers and existing classification methods are presented.

(5) The advantage of the testing-based classification model on handling outliers and controlling the Type I error rate in terms of FDR is empirically investigated. 

The rest of this paper is organized as follows. Section \ref{Related work} discusses some previous works that are related to our method. Section \ref{Methods} presents the details of our method. Section \ref{Experiments} reports experimental results on 40 real data sets. Section \ref{Relationship} discusses the relationship between our method and other approaches. Finally, Section \ref{Conclusion} concludes this paper.

\section{Related Work}
\label{Related work}
\subsection{Instance-based learning}
Instance-based learning is a lazy learning scheme in which the training instances are simply stored. When a new instance is encountered, a set of similar training instances are retrieved to classify the unknown testing instance. The most basic instance-based method is the \emph{k}-nearest neighbor algorithm (\emph{k}-NN) \citep{TITknn,Mitchell1997ML}, which assigns a new instance to the most common class among its \emph{k}-NNs in training instances. 

Essentially, our method can be considered as an instance-based learning approach since the two-sample test is conducted on the distance sets generated from all training instances or \emph{k}-NNs. This indicates that it is feasible to apply techniques developed for instance-based learning during the past decades \citep{wilson2000reduction,garcia2012prototype,derrac2014fuzzy} to further improve our method.

\subsection{Classification based on hypothesis testing}
Liao \& Akritas \citep{liao2007test} introduce a classification method based on hypothesis testing, which is abbreviated to TBC. Suppose there are two classes (positive vs. negative) in the training set, i.e., a binary classification problem, the issue is to allocate a new instance $t^*$ to one of the two classes. The basic idea of TBC is that if $t^*$ is placed into the wrong class, then the difference of two samples will be blurred. To implement this idea, two tests with respect to the equality of the means of two samples are conducted, in which $t^*$ is placed into the set of positive instances and the set of negative instances, respectively. Accordingly, we will obtain two \emph{p}-values $p_+$ and $p_{-}$, where $p_{+}$ ($p_{-}$) is generated from the test in which $t^*$ is assumed to belong to the positive (negative) class. If $p_{+}<p_{-}$, then $t^*$ is classified as a positive instance. Otherwise, $t^*$ will be classified as a negative instance. This method works well when the theoretical \emph{p}-values can be computed and compared. However, TBC has two problems. First, when the number of features of data set is larger than the sample size of one class, the \emph{p}-values cannot be computed at all because of the singularity of the sample covariance matrix. Second, when the instances from two class are well separated, the \emph{p}-values will equal to zero. 

Ghimire \& Wang \citep{ghimire2012classification} improve the TBC method by introducing a minimum distance into the method and come up with a new classifier for image pixels. Their new method works well in the context of image pixel classification.

Modarres \citep{modarres2014on,modarres2016multivariate,modarres2018multinomial} studies the properties of squared Euclidean interpoint distances (IPDs) between different samples which are taken from multivariate Bernoulli, multivariate Poisson and multinomial distributions. And he also discusses some applications based on IPDs within one sample and across two samples in different distributions. 

Afterwards, Guo \& Modarres \citep{guo2018interpoint} develop a classification method based on hypothesis testing, which is abbreviated to IDC. It is capable of classifying high dimensional instances by employing testing methods based on the IPDs between different instances. Several different test statistics based on IPDs have been discussed in \citep{guo2018interpoint} and we will take the Baringhaus and Franz (BF) statistic as the example. Given two sets of training instances, i.e., one positive set $D^{+}$ and one negative set $D^{-}$, IDC first computes the average IPDs within $D^{+}$, within $D^{-}$ and between $D^{+}$ and $D^{-}$, which are denoted by $\bar{d}_{D^{+}}$, $\bar{d}_{D^{-}}$ and $\bar{d}_{D^{+}D^{-}}$ respectively. Then, it calculates $BF_0=2\bar{d}_{D^{+}D^{-}}-\bar{d}_{D^{+}}-\bar{d}_{D^{-}}$. Similarly, $BF_1=2\bar{d}_{(D^{+} \cup \{t^*\})D^{-}}-\bar{d}_{D^{+} \cup \{t^*\}}-\bar{d}_{D^{-}}$ and $BF_2=2\bar{d}_{D^{+}(D^{-} \cup \{t^*\})}-\bar{d}_{D^{+}}-\bar{d}_{D^{-} \cup \{t^*\}}$ can be obtained by placing $t^*$ into $D^{+}$ and $D^{-}$, respectively. Note that $|BF_1-BF_0|$ ($|BF_2-BF_0|$) can be used to measure the change in the value of BF when $t^*$ is assigned to $D^{+}$ ($D^{-}$). Therefore, if $|BF_1-BF_0|<|BF_2-BF_0|$, $t^*$ is classified as a positive instance; otherwise, $t^*$ will be labelled as negative instance.

\subsection{Asymmetric classification error control}
In binary classification, most classifiers are constructed to minimize the overall classification error, which is a weighted sum of type I error (misclassifying a negative instance as a positive one) and type II error (misclassifying a positive instance as a negative one). However, in many realistic applications, different types of errors are often asymmetric, which have different costs and need to be treated with different weights. 

The cost-sensitive classification (CSC) method \citep{elkan2001foundations,zadrozny2003cost} can solve this problem to some extent. It takes the misclassification costs into consideration and aims to minimize the total cost of both errors. Another method is the Neyman-Pearson (NP) classification \citep{scott2005neyman}, which is inspired by classical NP hypothesis testing. It is a novel statistical framework for handling asymmetric type I/II error priorities and can seek a classifier that minimizes the type II error while maintaining the type I error below a user-specified level $\alpha$ \citep{tong2016A,tong2018neyman}. CSC and NP classification are fundamentally different approaches that have their own pros and cons \citep{scott2005neyman}. A main advantage of the NP classification is that it is a general framework that allows users to control type I classification error under $\alpha$ with a high probability.

It is very easy to control the type I error in terms of FDR in our formulation since the \emph{p}-values of each test instance with respect to different classes will be generated in the classification phase. In other words, such testing-based classification formulation provides a unified framework for controlling the asymmetric classification error in a natural way.

\section{Method}
\label{Methods}
\subsection{Two-sample testing}
Given two independent random samples $G_{X}$ and $G_{Y}$, where $G_{X}=\{x_{1},x_{2},...,x_{m}\}$ is drawn from the $X$ population and $G_{Y}=\{y_{1},y_{2},...,y_{n}\}$ is drawn from the $Y$ population, the general two-sample testing problem is concerned with the null hypothesis that the two samples are drawn from identical populations \citep{gibbons2011nonparametric}:\begin{displaymath}H_{0}:F_{X}(t) = F_{Y}(t)\; for\; all\; t, \end{displaymath} where $F_{X}$ and $F_{Y}$ are the cumulative distribution functions for the $X$ population and the $Y$ population, respectively.

% Problem formulation // 描述如何把分类转成testing的问题
\subsection{Problem formulation}
We consider the binary classification problem, in which the training set $D$ is composed of two disjoint sets $D^{+}$ and $D^{-}$. $D^{+}=\{t_{1}^{+},t_{2}^{+},..., t_{m}^{+}\}$ and $D^{-}=\{t_{1}^{-}, t_{2}^{-},..., t_{n}^{-}\}$ are called the positive training set and the negative training set, respectively. Given a test instance $t^{*}$, the classification task is to decide its class label (positive vs. negative).

We formulate the binary classification problem as a two-sample testing problem. In this formulation, the first sample $G_{X}$ is a set of \emph{m} observations, where the \emph{i}th observation is the distance between the test instance $t^{*}$ and the \emph{i}th training instance $t_{i}^{+}$ in $D^{+}$, i.e. $G_{X}=\{x_{i}|x_{i}=d(t^{*},t_{i}^{+}), 1\leq i\leq m\}$. Similarly, each observation in the second sample $G_{Y}$ is the distance between the test instance and each training instance in $D^{-}$, i.e. $G_{Y}=\{y_{j}|y_{j}=d(t^{*},t_{j}^{-}),1\leq j\leq n\}$. 

To conduct the standard classification task, we may test the null hypothesis against two alternative hypotheses $(F_{X}(t)<F_{Y}(t)$ and $F_{Y}(t)>F_{X}(t))$ to obtain two one-sided \emph{p}-values ($p_{X}$ and $p_{Y}$). If $p_{X}<p_{Y}$, we will label $t^{*}$ as a positive instance. Otherwise, we will classify $t^{*}$ as a negative instance.

To handle the multi-classification problem with $Q$ classes ($Q>2$), we can explore the one-vs-rest strategy by regarding the set of instances from one class as the positive training set and using the set of instances from the remaining classes as the negative training set. For each of $Q$ binary classification problems, we first conduct the two-sample testing to generate a one-sided \emph{p}-value for the corresponding class. Then, we can assign the test instance to the class that has the smallest \emph{p}-value.

\subsection{\emph{K}-NN variants}
In the above problem formulation, the distances to all training instances are utilized in the hypothesis testing. However, the existence of outlying and irrelevant training instances may decrease the classification accuracy. To alleviate this issue, we can conduct the hypothesis testing on two samples that are derived from the \emph{k}-NNs of the test instance. 

Under $H_{0}$, two natural \emph{k}-NN variants can be formulated. Similar to the \emph{k}-NN classifier, the first variant is to directly take the \emph{k}-NNs of the test instance to generate two samples. The distances from the test instance to these \emph{k} nearest training instances are divided into two groups according to the class label, where each group corresponds to one sample in our scenario. The second variant is to take $k_{1}$ nearest instances from $D^{+}$ and retrieve $k_{2}$ nearest instances from $D^{-}$ to generate two distance sets, where $\frac{k_{1}}{k_{2}}=\frac{m}{n}$. The rationale behind the second variant is that, if the null hypothesis is true, then the number of \emph{k}-NNs from each class is proportional to the number of training instances in that class. Since $k_{1}=k_{2}$ when $n=m$, we can take the same number of \emph{k}-NNs from each class in this case.

\subsection{The choice of testing methods}
The testing method for two-sample differences has been extensively investigated in the literature. One widely used test for this issue is the WMW test, which is also called the Mann-Whitney \emph{U} test or Wilcoxon rank-sum test \citep{Mann1947On}. To obtain the test statistic in WMW test, $G_{X}$ and $G_{Y}$ are merged to form a combined sample $G_{Z}=\{z_{1},z_{2},...,z_{m+n}\}$. Then, the observations in $G_{Z}$ are ordered:\begin{displaymath}z_{(1)}\leq z_{(2)}\leq ...\leq z_{(m+n)}.\end{displaymath} According to the ordered list, $R_{i1}$ is defined as the rank of $x_i$ in $G_{Z}$ and $R_{1}=\sum_{i=1}^{m}R_{i1}$. Then we can get $U_1=R_1-\frac{m(m+1)}{2}$. If the null hypothesis $H_{0}$ is true, then \begin{displaymath}Z=\frac{U_1-E(U_1|H_0)}{\sqrt{Var(U_1|H_0)}}\sim N(0,1),\end{displaymath}where\begin{displaymath}\begin{aligned}&E(U_1|H_0)=\frac{mn}{2},Var(U_1|H_0)=\frac{mn(m+n+1)}{12}.\end{aligned}\end{displaymath}Based on the above normal approximation, we can calculate the one-sided \emph{p}-value to test $H_{0}$ against $H_{1}$($F_{X}(t)<F_{Y}(t)$) for some $t$.
% Similarly, $R_{i2}$ denotes the rank of $y_i$ in $G_{Z}$ and $R_{2}=\sum_{i=1}^{n}R_{i2}$.

In our classification model, the choice of testing method is very flexible since the samples to be tested are unidimensional. That is, we can use any univariate two-sample testing method in our classifier. Therefore, we can also employ the testing methods such as pooled \emph{t}-test, two-sample Kolmogorov-Smirnov test \citep{wang2003evaluating} and precedence test instead of the WMW test. In Section \ref{Relationship}, we will further show that the use of different testing methods will establish the connection between our formulation and existing classification models.
 
\subsection{Handling outliers and FDR control}
As we have argued, the testing-based classification model has the advantage of controlling the FDR of classified test instances and handling outlying instances under the same framework. In general, we will assign the test instance to the class that has the smallest \emph{p}-value among \emph{Q} \emph{p}-values, where \emph{Q} is the number of classes. However, it is inappropriate to do so when all \emph{Q} \emph{p}-values are not significant. Luckily, we can use FDR \citep{Benjamini1995Controlling} to tackle this problem. We can obtain \emph{Q} sets of \emph{p}-values from all test instances because our method returns \emph{Q} \emph{p}-values to classify every test instance. Every \emph{p}-value set is firstly sorted in a non-descending order: $p_1 \leq p_2 \leq ... \leq p_{u}$, where $u$ is the number of all test instances. Given a significance level $\alpha$, let $i_{max}$ be the largest index for which \begin{displaymath}p_{i}\leq \frac{i \times \alpha}{u}.\end{displaymath} If $i \leq i_{max}$, then the corresponding test instance will be assigned to the current class. After conducting FDR control on all \emph{Q} \emph{p}-value sets, we can label the test instances that are not classified to any class as outliers.

\section{Experiments}
\label{Experiments}
\begin{table}\centering
\caption{The detailed characteristics of the forty data sets. For each data set, the number of instances without (with) missing values is provided outside (inside) the parentheses in the second column. The class distribution information, i.e. the number of instances in every class, is given in the 5th column. The last column provides links to download the corresponding data set.}
\begin{tabular}{llcccccc} \hline
ID & Names & Instances & Features & Classes & Download Links \\ \hline
1 & Appendicitis & 106 & 7 & 2 & \href{http://sci2s.ugr.es/keel/dataset.php?cod=183}{KEEL} \\
2 & Balance & 625 & 4 & 3 & \href{http://archive.ics.uci.edu/ml/datasets/Balance+Scale}{UCI}, \href{http://sci2s.ugr.es/keel/dataset.php?cod=54}{KEEL} \\
3 & Banana & 5300 & 2 & 2 & 
\href{http://sci2s.ugr.es/keel/dataset.php?cod=182}{KEEL} \\
4 & Bands & 365(539) & 19 & 2 & \href{http://archive.ics.uci.edu/ml/datasets/Cylinder+Bands}{UCI}, \href{http://sci2s.ugr.es/keel/dataset.php?cod=89}{KEEL} \\
5 & Bupa & 345 & 6 & 2 & \href{http://archive.ics.uci.edu/ml/datasets/Liver+Disorders}{UCI}, \href{http://sci2s.ugr.es/keel/dataset.php?cod=55}{KEEL} \\
6 & Cleveland & 297(303) & 13 & 5 & \href{http://archive.ics.uci.edu/ml/datasets/Heart+Disease}{UCI}, \href{http://sci2s.ugr.es/keel/dataset.php?cod=57}{KEEL} \\
7 & Dermatology & 358(366) & 34 & 6 & \href{http://archive.ics.uci.edu/ml/datasets/Dermatology}{UCI}, \href{http://sci2s.ugr.es/keel/dataset.php?cod=60}{KEEL} \\
8 & Haberman & 306 & 3 & 2 & \href{http://archive.ics.uci.edu/ml/datasets/Haberman\%27s+Survival}{UCI}, \href{http://sci2s.ugr.es/keel/dataset.php?cod=62}{KEEL} \\
9 & Hayes-roth & 160 & 4 & 3 & \href{http://archive.ics.uci.edu/ml/datasets/Hayes-Roth}{UCI}, \href{http://sci2s.ugr.es/keel/dataset.php?cod=186}{KEEL} \\
10 & Heart & 270 & 13 & 2 & \href{http://archive.ics.uci.edu/ml/datasets/Statlog+\%28Heart\%29}{UCI}, \href{http://sci2s.ugr.es/keel/dataset.php?cod=99}{KEEL} \\
11 & Hepatitis & 80(155) & 19 & 2 & \href{http://archive.ics.uci.edu/ml/datasets/Hepatitis}{UCI}, \href{http://sci2s.ugr.es/keel/dataset.php?cod=100}{KEEL} \\
12 & Ionosphere & 351 & 34 & 2 & \href{http://archive.ics.uci.edu/ml/datasets/Ionosphere}{UCI}, \href{http://sci2s.ugr.es/keel/dataset.php?cod=101}{KEEL} \\
13 & Iris & 150 & 4 & 3 & \href{http://archive.ics.uci.edu/ml/datasets/Iris}{UCI}, \href{http://sci2s.ugr.es/keel/dataset.php?cod=18}{KEEL} \\
14 & Led7digit & 500 & 7 & 10 & \href{http://archive.ics.uci.edu/ml/datasets/LED+Display+Domain}{UCI}, \href{http://sci2s.ugr.es/keel/dataset.php?cod=63}{KEEL} \\
15 & Mammographic & 830(961) & 5 & 2 & \href{http://archive.ics.uci.edu/ml/datasets/Mammographic+Mass}{UCI}, \href{http://sci2s.ugr.es/keel/dataset.php?cod=86}{KEEL} \\
16 & Marketing & 6876(8993) & 13 & 9 & \href{http://sci2s.ugr.es/keel/dataset.php?cod=163}{KEEL} \\
17 & Monks-2 & 432 & 7 & 2 & \href{http://archive.ics.uci.edu/ml/datasets/MONK\%27s+Problems}{UCI}, \href{http://sci2s.ugr.es/keel/dataset.php?cod=65}{KEEL} \\
18 & Movement\_libras & 360 & 90 & 15 & \href{http://archive.ics.uci.edu/ml/datasets/Libras+Movement}{UCI}, \href{http://sci2s.ugr.es/keel/dataset.php?cod=165}{KEEL} \\
19 & Newthyroid & 215 & 5 & 3 & \href{http://archive.ics.uci.edu/ml/datasets/Thyroid+Disease}{UCI}, \href{http://sci2s.ugr.es/keel/dataset.php?cod=66}{KEEL} \\
20 & Page-blocks & 5473 & 10 & 5 & \href{http://archive.ics.uci.edu/ml/datasets/Page+Blocks+Classification}{UCI}, \href{http://sci2s.ugr.es/keel/dataset.php?cod=104}{KEEL} \\
21 & Penbased & 10092 & 16 & 10 & \href{http://archive.ics.uci.edu/ml/datasets/Pen-Based+Recognition+of+Handwritten+Digits}{UCI}, \href{http://sci2s.ugr.es/keel/dataset.php?cod=70}{KEEL} \\
22 & Phoneme & 5404 & 5 & 2 & \href{https://www.elen.ucl.ac.be/neural-nets/Research/Projects/ELENA/databases/REAL/phoneme/}{UCL}, \href{http://sci2s.ugr.es/keel/dataset.php?cod=105}{KEEL} \\
23 & Pima & 768 & 8 & 2 & \href{http://archive.ics.uci.edu/ml/datasets/Pima+Indians+Diabetes}{UCI}, \href{http://sci2s.ugr.es/keel/dataset.php?cod=21}{KEEL} \\
24 & Ring & 7400 & 20 & 2 & \href{http://www.cs.utoronto.ca/~delve/data/ringnorm/desc.html}{TORONTO}, \href{http://sci2s.ugr.es/keel/dataset.php?cod=106}{KEEL} \\
25 & Satimage & 6435 & 36 & 7 & \href{http://archive.ics.uci.edu/ml/datasets/Statlog+\%28Landsat+Satellite\%29}{UCI}, \href{http://sci2s.ugr.es/keel/dataset.php?cod=71}{KEEL} \\
26 & Segment & 2310 & 19 & 7 & \href{http://archive.ics.uci.edu/ml/datasets/Image+Segmentation}{UCI}, \href{http://sci2s.ugr.es/keel/dataset.php?cod=107}{KEEL} \\
27 & Sonar & 208 & 60 & 2 & \href{http://archive.ics.uci.edu/ml/datasets/Connectionist+Bench+\%28Sonar\%2C+Mines+vs.+Rocks\%29}{UCI}, \href{http://sci2s.ugr.es/keel/dataset.php?cod=85}{KEEL} \\
28 & Spambase & 4597(4601) & 57 & 2 & \href{http://archive.ics.uci.edu/ml/datasets/Spambase}{UCI}, \href{http://sci2s.ugr.es/keel/dataset.php?cod=109}{KEEL} \\
29 & Spectfheart & 267 & 44 & 2 & \href{http://archive.ics.uci.edu/ml/datasets/SPECTF+Heart}{UCI}, \href{http://sci2s.ugr.es/keel/dataset.php?cod=185}{KEEL} \\
30 & Tae & 151 & 5 & 3 & \href{http://archive.ics.uci.edu/ml/datasets/Teaching+Assistant+Evaluation}{UCI}, \href{http://sci2s.ugr.es/keel/dataset.php?cod=188}{KEEL} \\
31 & Texture & 5500 & 40 & 11 & \href{https://www.elen.ucl.ac.be/neural-nets/Research/Projects/ELENA/databases/REAL/texture}{UCL}, \href{http://sci2s.ugr.es/keel/dataset.php?cod=72}{KEEL} \\
32 & Thyroid & 7200 & 21 & 3 & \href{http://archive.ics.uci.edu/ml/datasets/Thyroid+Disease}{UCI}, \href{http://sci2s.ugr.es/keel/dataset.php?cod=67}{KEEL} \\
33 & Titanic & 2201 & 3 & 2 & \href{http://www.cs.toronto.edu/~delve/data/titanic/titanicDetail.html}{TORONTO}, \href{http://sci2s.ugr.es/keel/dataset.php?cod=189}{KEEL} \\
34 & Twonorm & 7400 & 20 & 2 & \href{http://www.cs.utoronto.ca/~delve/data/twonorm/desc.html}{
TORONTO}, \href{http://sci2s.ugr.es/keel/dataset.php?cod=110}{KEEL} \\
35 & Vehicle & 846 & 18 & 4 & \href{http://archive.ics.uci.edu/ml/datasets/Statlog+\%28Vehicle+Silhouettes\%29}{UCI}, \href{http://sci2s.ugr.es/keel/dataset.php?cod=68}{KEEL} \\
36 & Vowel & 990 & 13 & 11 & \href{http://archive.ics.uci.edu/ml/datasets/Connectionist+Bench+\%28Vowel+Recognition+-+Deterding+Data\%29}{UCI}, \href{http://sci2s.ugr.es/keel/dataset.php?cod=113}{KEEL} \\
37 & Wdbc & 569 & 30 & 2 & \href{http://archive.ics.uci.edu/ml/datasets/Breast+Cancer+Wisconsin+\%28Diagnostic\%29}{UCI}, \href{http://sci2s.ugr.es/keel/dataset.php?cod=111}{KEEL} \\
38 & Wine & 178 & 13 & 3 & \href{http://archive.ics.uci.edu/ml/datasets/Wine}{UCI}, \href{http://sci2s.ugr.es/keel/dataset.php?cod=31}{KEEL} \\
39 & Winequality-red & 1599 & 11 & 6 & \href{http://archive.ics.uci.edu/ml/datasets/Wine+Quality}{UCI}, \href{http://sci2s.ugr.es/keel/dataset.php?cod=210}{KEEL} \\
40 & Wisconsin & 683(699) & 9 & 2 & \href{http://archive.ics.uci.edu/ml/datasets/Breast+Cancer+Wisconsin+\%28Original\%29}{UCI}, \href{http://sci2s.ugr.es/keel/dataset.php?cod=73}{KEEL} \\
\hline
\end{tabular}
\label{table1}
\end{table}

\subsection{Data sets and experimental settings}
We have conducted experiments on 40 data sets from the UCI \citep{Dua2017} repository and the KEEL-dataset repository \citep{Alcal2011KEEL}. Among these data sets, the number of instances ranges from 80 to 10092 and the number of features varies from 2 to 90. Most data sets have less than 10 classes and only six of them have more than 10 classes. The detailed characteristics of these data sets are given in Table \ref{table1}. Moreover, the instances with missing values are discarded and the numeric feature values are normalized into the interval $[0,1]$ in the pre-processing process.

In the experiment, we perform 10-fold cross-validation (CV) and count the number of instances which have been correctly classified to compute a classification accuracy value. For every data set, we repeat the 10-fold CV experiment 10 times and record the average and standard deviation of 10 accuracy values as the final results.

\begin{table}\centering
\caption{The average accuracy over forty data sets for IBT-U and IBT-U-K variants (\emph{k}=3).}
\begin{tabular}{lc}
\hline
Methods  &  Avg accuracy\\ \hline
IBT-U   &  0.6795  \\
IBT-U-K-D  &  0.8027  \\
IBT-U-K-S  &  0.7906  \\ \hline
\end{tabular}
\label{table2}
\end{table}

\begin{table}\centering
\caption{The average accuracy over forty data sets for two IBT-U-K variants.}
\begin{tabular}{lcccc}
\hline
Methods  & \emph{k}=3 & \emph{k}=5 & \emph{k}=7 & \emph{k}=9 \\  \hline
IBT-U-K-D & 0.8027 & 0.7835 & 0.7677 & 0.7547 \\
IBT-U-K-S & 0.7906 & 0.7829 & 0.7742 & 0.7703 \\ \hline
\end{tabular}
\label{table3}
\end{table}

\subsection{All instances vs. \emph{k}-NNs}
In the first experiment, we compare several variants of our formulation to check which one is better in practice. Since our method is a classifier that combines instance-based learning and hypothesis testing, we will use the abbreviation IBT to denote such a classification model. To distinguish different variants, IBT-U is used to denote the classification model when the Mann-Whitney \emph{U} test is applied to the distance sets derived from all training instances. Similarly, IBT-U-K is used to denote the classification model in which the distance sets are generated according to \emph{k}-NNs of the test instance. Furthermore, two \emph{k}-NN variants are denoted by IBT-U-K-D (\emph{k}-NNs are obtained \textbf{D}irectly without considering the class label) and IBT-U-K-S (\emph{k}-NNs are obtained \textbf{S}eparately from different classes), respectively.

Additionally, the parameter \emph{k} for two \emph{k}-NN variants is specified as 3,5,7 and 9, respectively. The detailed experimental results on these three variants are given in Appendix Table \ref{table6}, Appendix Table \ref{table7} and Appendix Table \ref{table8} and their average accuracies are summarized in Table \ref{table2} and Table \ref{table3}.

As shown in Table \ref{table2}, the performance of IBT-U is much worse than that of two \emph{k}-NN variants. This indicates that it is plausible to explore the \emph{k}-NN strategy in the testing-based classification model. As shown in Table \ref{table3}, the average classification accuracies of two \emph{k}-NN variants are quite similar when \emph{k} is varied from 3 to 9. In the forthcoming sections, we will use IBT-U-K-D (\emph{k}=3) as a representative of our classifiers in the performance comparison.

\subsection{Our method vs. Other testing-based classifiers}
In the second experiment, we compare our method with two previous methods, TBC \citep{liao2007test} and IDC \citep{guo2018interpoint}, which also use hypothesis testing to solve a classification problem. The detailed experimental results are given in Appendix Table \ref{table9} and their average accuracies are presented in Table \ref{table4}.

In the implementation of TBC, we employ the Hotelling's ${T^2}$ test as the testing method, which has been utilized in \citep{liao2007test}. And we use the Hotelling's ${T^2}$ statistics instead of \emph{p}-values in the classification since the generated \emph{p}-values are often zeros. In the implementation of IDC, we use the Baringhaus and Franz (BF) statistic as the test statistic and assume equal prior probabilities in splite of unequal sample sizes.

For TBC, the classification accuracies on five data sets (Cleveland, Dermatology, Hepatitis, Movement\_libras and Winequality-red) are 0 because the number of features of these data sets is larger than the sample size of one class, so we only use the rest 35 data sets to compute the average classification accuracy. For IDC, it can be applied to all data sets, so we simply compute the average of 40 accuracy values. According the comparison result, it's obvious to see that our method performs significantly better than TBC and IDC. 

Among these three methods, our method can achieve the best performance due to the following reasons. First, our method only consider the \emph{k}-NNs of test instance while TBC and IDC utilize all training instances without considering the existence of outlying and irrelevent ones. Second, our method employs a hypothesis testing strategy that is totally different from that used in TBC and IDC. 

\begin{table}\centering
\caption{The average accuracy for three testing-based classification methods: TBC, IDC and our method (IBT-U-K-D, \emph{k}=3).}
\begin{tabular}{lc}
\hline
Methods  &  Avg accuracy\\ \hline
TBC  &  0.5901  \\
IDC  &  0.6859  \\
Our method  &  0.8027 \\ \hline
\end{tabular}
\label{table4}
\end{table}

\begin{table}\centering
\caption{The average accuracy for three classic classifiers: \emph{k}-NN (\emph{k}=3), SVM, decision tree (DT) and our method (IBT-U-K-D, \emph{k}=3).}
\begin{tabular}{lc}
\hline
Methods  &  Avg accuracy\\ \hline
\emph{k}-NN  &  0.8058  \\
SVM  &  0.7928  \\
DT   &  0.8003  \\
Our method  &  0.8027 \\ \hline
\end{tabular}
\label{table5}
\end{table}

\subsection{Our method vs. Classic classifiers}
In the third experiment, we compare our method with three classic classifiers: \emph{k}-NN, support vector machine (SVM) and decision tree (DT). The detailed experimental results are given in Appendix Table \ref{table10} and Appendix Table \ref{table11} and their average accuracies are presented in Table \ref{table5}.

For SVM, \emph{k}-NN and DT, we use the functions \emph{fitcecoc}, \emph{fitcknn} and \emph{fitctree} with their default parameter settings in Matlab 2018b, respectively. The reason for using \emph{fitcecoc} function is that it can generate a multi-class model for SVM.

As shown in Table \ref{table5}, our method is able to achieve the same level performance as these classic classifiers. Concretely, there are 13, 19 and 18 data sets on which our method can produce higher classification accuracies than \emph{k}-NN, SVM and DT among the 40 data sets, respectively. In a word, our method is competitive to these classic classifiers with respect to the overall performance.

\subsection{Handling outliers through FDR control}
In the last experiment, we investigate the potential of our method on outlier detection and FDR control. The \emph{balance} data set from UCI is used as an example, which has 625 instances and three classes (\emph{L}, \emph{B} and \emph{R}). There are 288, 49 and 288 instances in the three classes respectively, as shown in Table 5. If we take a subset of the 576 (288+288) instances from the class \emph{L} and \emph{R} as training instances and use the 49 instances from the class \emph{B} as test instances, then it is obvious that all test instances should be considered as outliers.

We randomly take 80 percent of instances from the class \emph{L} and \emph{R} to compose the training set. In order to obtain the average performance, 10 different random training sets are generated. We use IBT-U as the classifier and the significance level for FDR is set to be 0.05. The experimental results show that 48 of 49 test instances can be labelled as outliers on average. Specifically, there are at most two test instances which cannot be labelled as outliers and they are usually different when the training set is different. Therefore, our method is able to recognize outliers and control the FDR of classification results in the same time.

\section{Relationship to Other Approaches}
\label{Relationship}
Our classification method is a two-phase approach: two distance sets are first generated and then the two-sample test is conducted. As we have discussed, we may use different significance testing methods in the second phase. In this section, we will show that the use of different testing methods will lead to different classifiers that have close relationship with existing classification models.

\subsection{Connection to Nearest Centroid Classifier}
The nearest centroid (mean) classifier is one of the most widely used instance-based classification models \citep{friedman2001elements}. In the training phase, only the centroid for each class is calculated and stored. In the classification phase, the distance between one unknown instance and each centroid is calculated to find the nearest centroid. Then, this new test instance is assigned to the class of its nearest centroid. 

If the pooled \emph{t}-test is employed as the significance testing procedure in our model, then we can reveal some interesting connections between our method and the nearest centroid classifier. To simplify the analysis, we first consider the scenario of univariate data set and then discuss the case of multivariate data set.

Given two one-dimensional sets $D^{+}=\{t_{1}^{+},t_{2}^{+},..., t_{m}^{+}\}$ and $D^{-}=\{t_{1}^{-},t_{2}^{-},..., t_{n}^{-}\}$, their centroids (means) can be easily computed by $C_{D^+}=\frac{1}{m}\sum_{i=1}^{m}{t^+_{i}}$ and $C_{D^-}=\frac{1}{n}\sum_{j=1}^{n}{t^-_{j}}$. Given an unknown instance $t^*$, the distances between $t^*$ and these two centroids can be measured by $d^+=|t^*-C_{D^+}|$ and $d^-=|t^*-C_{D^-}|$. The nearest centroid classification method will assign $t^*$ to the positive or the negative class according to whether $d^+<d^-$.

In our method, two samples $G_{X}=\{|t^{*}-t_{i}^{+}|, 1\leq i\leq m\}$ and $G_{Y}=\{|t^{*}-t_{j}^{+}|, 1\leq j\leq n\}$ are obtained and their means are denoted by $\bar{d}_X=\frac{1}{m}\sum_{i=1}^{m}|t^*-t_{i}^+|$ and $\bar{d}_Y=\frac{1}{n}\sum_{j=1}^{n}|t^*-t_{j}^-|$. Then, we test the null hypothesis against two alternative hypotheses $(F_{X}(t)<F_{Y}(t)$ and $F_{Y}(t)>F_{X}(t))$ on the two samples to obtain two one-sided \emph{p}-values ($p_{X}$ and $p_{Y}$). At last, our method will assign $t^*$ to the positive (negative) class if $p_{X}<p_{Y}$ ($p_{X}>p_{Y}$). 

Note that when the pooled \emph{t}-test is employed in our method, we will obtain two \emph{t} statistics ($t_{X}$ and $t_{Y}$). We can get \begin{align*}p_{X}<p_{Y} & \Leftrightarrow t_X < t_Y \\ & \Leftrightarrow \bar{d}_X-\bar{d}_Y<\bar{d}_Y-\bar{d}_X \\ & \Leftrightarrow \bar{d}_X < \bar{d}_Y.\end{align*} Similarly, we can also get $p_{X}>p_{Y} \Leftrightarrow \bar{d}_X > \bar{d}_Y$. Therefore, our method will assign $t^*$ to the positive class if $\bar{d}_X < \bar{d}_Y$. Otherwise, we will label $t^*$ as a negative instance. 

According to the triangle inequality, we can get \begin{align*}d^+ &= |t^*-C_{D^+}|\\ &=|t^*-\frac{1}{m}\sum_{i=1}^{m}{t^+_{i}}|\\ &= \frac{1}{m}|mt^*-\sum_{i=1}^{m}t_{i}^+|\\ &\leq \frac{1}{m}\sum_{i=1}^{m}|t^*- t_{i}^+| \\ &= \bar{d}_X\end{align*} in which the equality holds if and only if $t^* \geq \underset{1 \leq i\leq m}{max}{t_{i}^{+}}$ or $t^* \leq \underset{1 \leq i\leq m}{min}{t_{i}^{+}}$. Similarly, we can get $d^- \leq \bar{d}_Y$ in which the equality holds if and only if $t^* \geq \underset{1 \leq i\leq m}{max}{t_{i}^{-}}$ or $t^* \leq \underset{1 \leq i\leq m}{min}{t_{i}^{-}}$.

When $d^+=\bar{d}_X$ and $d^-=\bar{d}_Y$, our method will assign the test instance to the same class label as the nearest centroid classification method. Obviously, the above analysis establish the equivalence between our method and the nearest centroid classifier under very strict constraints: (1) one-dimensional data set, (2) the test instance is no less (more) than all training instances in each class.

For the multivariate case, it is very difficult to analyze their relationship in a quantitative manner. One naive connection is that if $(d_{X}-d_{Y})(d^+-d^-)>0$, then our method and the nearest centroid classification method will produce the same classification result.

\subsection{Connection to \emph{k}-NN Classifier}
The \emph{k}-NN classifier is one of the most popular classification methods in the literature \citep{wu2008top}. In our formulation, if the precedence test \citep{gibbons2011nonparametric} is employed as the significance testing method, then we may uncover some interesting connections between our method and the \emph{k}-NN classifier. 

We still consider the binary classification problem in which the training data is composed of $m$ positive instances from $D^+$ and $n$ negative instances from $D^-$. Given an unknown instance $t^*$, the \emph{k}-NN classification method finds its \emph{k} nearest neighbors (\emph{k}-NNs) to conduct the classification. These \emph{k}-NNs can be divided into two groups: $k^+$ positive instances from $D^+$ and $k^{-}$ instances from $D^{-}$, where $k = k^{+}+k^{-}$. If $k^{+}>k^{-}$, then $t^*$ will be classified as a positive instance. Otherwise, $t^*$ is assigned to the negative class.

The precedence test is a two-sample test based on the order of early failures \citep{balakrishnan2006precedence}. Given two independent samples, $G_{X} = \{x_1, x_2,..., x_m\}$ and $G_{Y}=\{y_1, y_2, ...., y_n\}$, let $x_{(1)} \leq x_{(2)} \leq ... \leq x_{(m)}$ and $y_{(1)} \leq y_{(2)} \leq ... \leq y_{(n)}$ denote their order statistics. The precedence test is based on the number of observations from one sample which exceed (precede) some threshold specified by the other sample. More precisely, the test statistic $W_{r}$ is the number of observations in $G_{X}$ that precede the \emph{r}-th order statistic $y_{(r)}$ from $G_{Y}$. Alternatively, one can use the number of observations in $G_{Y}$ that exceed the \emph{s}-th order statistic $x_{(s)}$ from $G_{x}$ as the test statistic $W_s$. Large values of these two test statistics will lead to the rejection of the null hypothesis that two distributions are equal.

In our problem formulation, $G_{X}$ ($G_{Y}$) is the distance set between $t^*$ and the instances in $D^+$ ($D^-$). Then, $x_{(1)}, x_{(2)}, ..., x_{(k^+)}, y_{(1)}, y_{(2)},...,y_{(k^{-})}$ will be the \emph{k} distance values between $t^*$ and its $k$-NNs. If we use the precedence test as the significance testing method and suppose that $x_{(k^{+})}\leq y_{(k^{-}+1)} \leq x_{(k^{+}+1)}$, we can set $r=k^{-}+1$ to obtain the corresponding test statistic $W_{r} = k^{+}$ for testing the null hypothesis against the alternative hypothesis ($F_X<F_Y$). Alternatively, if we let $s = k^{+}+1$, we can obtain another test statistic $W_{s} = k^{-}$ for testing the null hypothesis against the alternative hypothesis ($F_X>F_Y$). And we can also get two \emph{p}-values, $p_{X}$ and $p_{Y}$. At last, $t^*$ will be assigned to the positive (negative) class if the former (latter) is smaller.

If we further assume that the positive training set and the negative training set have the same size, i.e., $m=n$, then the two \emph{p}-values will be totally determined by the two test statistics: $p_{X}<p_{Y} \Leftrightarrow k^{+}> k^{-}$ or $p_{X}>p_{Y} \Leftrightarrow k^{+}< k^{-}$. Therefore, our method and the \emph{k}-NN classifier will generate the same classification result under the above assumptions. From this aspect, we may regard our method equipped with the precedence test as a generalized ''statistical'' \emph{k}-NN classifier.

\section{Conclusion}
\label{Conclusion}
Due to the importance of the classification problem, many effective classification algorithms have been proposed from different societies. However, most work on classification does not address the issue of statistical significance. Towards this direction, several initial research efforts have investigated the feasibility of constructing a classifier through significance testing. Unfortunately, this interesting idea has not receive much attention during the past 10 years. This is mainly because the following reasons: (1) there are still no such testing-based classifiers that can achieve the same level performance as the state-of-the-art methods on real data sets; (2) the potential benefit of deploying such testing-based classifiers is still not clear. 

Based on the above observations, this paper takes one step further towards this direction by formulating the classification problem as a two-sample testing problem. This new formulation enables us to generate several testing-based classifiers that have comparable performance with standard classifiers such as SVM. In addition, we show that it is quite easy to handle outlying test instances and control the FDR of classification results based on the \emph{p}-values associated with each test instance. 

We believe this paper will significantly contribute to the development of testing-based classification model, which will become a new promising classifier family. As the study on the testing-based classification model is still in its infancy stage, many research issues remain unexplored and should be further investigated in the future work. For example, since all the existing testing-based classifiers are based on the idea of instance-based learning, how to build a non-lazy testing-based classifier will be an interesting and challenging issue.

%
% The acknowledgments section is defined using the "acks" environment (and NOT an unnumbered section). This ensures
% the proper identification of the section in the article metadata, and the consistent spelling of the heading.
\begin{acks}
This work was partially supported by the Natural Science Foundation of China (Nos. 61572094, 61771331) and the Fundamental Research Funds for the Central Universities (No. DUT2017TB02).
\end{acks}

%
% The next two lines define the bibliography style to be used, and the bibliography file.
\bibliographystyle{ACM-Reference-Format}
\bibliography{ref}

% 
% If your work has an appendix, this is the place to put it.
\appendix

\section{Detailed Experimental Results}

\subsection{}

The detailed experimental results of IBT-U are given by Table 6.
\begin{table}\centering
\caption{The detailed experimental results of IBT-U.}
\begin{tabular}{llcc}
\hline
ID & Names & Avg & Std \\ \hline
1 & Appendicitis & 0.8557 & 0.0046 \\
2 & Balance & 0.8800 & 0.0039 \\
3 & Banana & 0.5998 & 0.0017 \\
4 & Bands & 0.6405 & 0.0128 \\
5 & Bupa & 0.5574 & 0.0170 \\
6 & Cleveland & 0.5505 & 0.0048 \\
7 & Dermatology & 0.8944 & 0.0041 \\
8 & Haberman & 0.7144 & 0.0166 \\
9 & Hayes-roth & 0.5581 & 0.0221 \\
10 & Heart & 0.8241 & 0.0047 \\
11 & Hepatitis & 0.8088 & 0.0084 \\
12 & Ionosphere & 0.6638 & 0.0033 \\
13 & Iris & 0.9567 & 0.0047 \\
14 & Led7digit & 0.7206 & 0.0076 \\
15 & Mammographic & 0.7952 & 0.0000 \\
16 & Marketing & 0.2995 & 0.0015 \\
17 & Monks-2 & 0.5185 & 0.0149 \\
18 & Movement\_libras & 0.3883 & 0.0146 \\
19 & Newthyroid & 0.8581 & 0.0025 \\
20 & Page-blocks & 0.9043 & 0.0005 \\
21 & Penbased & 0.5566 & 0.0005 \\
22 & Phoneme & 0.7172 & 0.0008 \\
23 & Pima & 0.7233 & 0.0032 \\
24 & Ring & 0.5049 & 0.0000 \\
25 & Satimage & 0.7262 & 0.0005 \\
26 & Segment & 0.7923 & 0.0013 \\
27 & Sonar & 0.6861 & 0.0204 \\
28 & Spambase & 0.8241 & 0.0008 \\
29 & Spectfheart & 0.4097 & 0.0054 \\
30 & Tae & 0.3861 & 0.0125 \\
31 & Texture & 0.7414 & 0.0009 \\
32 & Thyroid & 0.3158 & 0.0015 \\
33 & Titanic & 0.7760 & 0.0000 \\
34 & Twonorm & 0.9770 & 0.0003 \\
35 & Vehicle & 0.4375 & 0.0086 \\
36 & Vowel & 0.2748 & 0.0060 \\
37 & Wdbc & 0.9404 & 0.0010 \\
38 & Wine & 0.9416 & 0.0039 \\
39 & Winequality-red & 0.5131 & 0.0035 \\
40 & Wisconsin & 0.9458 & 0.0000 \\ \hline
& Avg & 0.6795 & 0.0055 \\ \hline
\end{tabular}
\label{table6}
\end{table}

\subsection{}

The detailed experimental results of IBT-U-K-D are given by Table 7.
\begin{table}\centering
\caption{The detailed experimental results of IBT-U-K-D.}
\begin{tabular}{llcccccccc}
\hline
\multirow{2}{*}{ID} & \multirow{2}{*}{Names} & \multicolumn{2}{c}{\emph{k}=3} & \multicolumn{2}{c}{\emph{k}=5} & \multicolumn{2}{c}{\emph{k}=7} & \multicolumn{2}{c}{\emph{k}=9}  \\ \cline{3-10} 
              &        & \multicolumn{1}{c}{Avg} & \multicolumn{1}{c}{Std} & \multicolumn{1}{c}{Avg} & \multicolumn{1}{c}{Std} & \multicolumn{1}{c}{Avg} & \multicolumn{1}{c}{Std} & \multicolumn{1}{c}{Avg} & \multicolumn{1}{c}{Std} \\ \hline
1 & Appendicitis & 0.8283 & 0.0116 & 0.7764 & 0.0141 & 0.7642 & 0.0252 & 0.7170 & 0.0209 \\
2 & Balance & 0.7782 & 0.0030 & 0.7528 & 0.0039 & 0.7184 & 0.0065 & 0.6834 & 0.0078 \\
3 & Banana & 0.8642 & 0.0016 & 0.8500 & 0.0020 & 0.8338 & 0.0024 & 0.8238 & 0.0013 \\
4 & Bands & 0.6978 & 0.0132 & 0.6726 & 0.0147 & 0.6564 & 0.0121 & 0.6452 & 0.0226 \\
5 & Bupa & 0.5986 & 0.0087 & 0.5948 & 0.0116 & 0.5797 & 0.0196 & 0.5713 & 0.0119 \\
6 & Cleveland & 0.5380 & 0.0074 & 0.5091 & 0.0191 & 0.4707 & 0.0122 & 0.4609 & 0.0150 \\
7 & Dermatology & 0.9402 & 0.0046 & 0.9349 & 0.0076 & 0.9179 & 0.0106 & 0.9101 & 0.0072 \\
8 & Haberman & 0.6585 & 0.0118 & 0.6585 & 0.0180 & 0.6261 & 0.0135 & 0.5971 & 0.0096 \\
9 & Hayes-roth & 0.7500 & 0.0189 & 0.7256 & 0.0163 & 0.7038 & 0.0232 & 0.6969 & 0.0221 \\
10 & Heart & 0.7552 & 0.0088 & 0.6856 & 0.0126 & 0.6722 & 0.0142 & 0.6652 & 0.0160 \\
11 & Hepatitis & 0.8150 & 0.0115 & 0.7850 & 0.0287 & 0.7425 & 0.0251 & 0.7363 & 0.0161 \\
12 & Ionosphere & 0.8556 & 0.0052 & 0.8575 & 0.0059 & 0.8558 & 0.0043 & 0.8541 & 0.0078 \\
13 & Iris & 0.9600 & 0.0054 & 0.9420 & 0.0077 & 0.9053 & 0.0129 & 0.9127 & 0.0097 \\
14 & Led7digit & 0.5770 & 0.0091 & 0.5230 & 0.0162 & 0.4604 & 0.0089 & 0.4286 & 0.0072 \\
15 & Mammographic & 0.7171 & 0.0061 & 0.7045 & 0.0084 & 0.6745 & 0.0069 & 0.6508 & 0.0055 \\
16 & Marketing & 0.2573 & 0.0016 & 0.2567 & 0.0025 & 0.2553 & 0.0024 & 0.2480 & 0.0027 \\
17 & Monks-2 & 0.7704 & 0.0124 & 0.7683 & 0.0174 & 0.7745 & 0.0182 & 0.7745 & 0.0170 \\
18 & Movement\_libras & 0.8181 & 0.0086 & 0.8036 & 0.0113 & 0.7978 & 0.0084 & 0.7875 & 0.0155 \\
19 & Newthyroid & 0.9614 & 0.0058 & 0.9581 & 0.0062 & 0.9470 & 0.0070 & 0.9474 & 0.0054 \\
20 & Page-blocks & 0.9534 & 0.0013 & 0.9466 & 0.0015 & 0.9405 & 0.0013 & 0.9361 & 0.0016 \\
21 & Penbased & 0.9931 & 0.0002 & 0.9915 & 0.0005 & 0.9896 & 0.0004 & 0.9876 & 0.0005 \\
22 & Phoneme & 0.8900 & 0.0014 & 0.8675 & 0.0022 & 0.8516 & 0.0020 & 0.8415 & 0.0033 \\
23 & Pima & 0.6915 & 0.0089 & 0.6634 & 0.0096 & 0.6406 & 0.0135 & 0.6319 & 0.0134 \\
24 & Ring & 0.7894 & 0.0013 & 0.7948 & 0.0016 & 0.8003 & 0.0020 & 0.8041 & 0.0018 \\
25 & Satimage & 0.8949 & 0.0012 & 0.8827 & 0.0027 & 0.8706 & 0.0022 & 0.8634 & 0.0022 \\
26 & Segment & 0.9640 & 0.0017 & 0.9572 & 0.0017 & 0.9513 & 0.0017 & 0.9396 & 0.0027 \\
27 & Sonar & 0.8630 & 0.0089 & 0.8452 & 0.0115 & 0.8260 & 0.0084 & 0.7957 & 0.0109 \\
28 & Spambase & 0.8978 & 0.0017 & 0.8704 & 0.0021 & 0.8458 & 0.0026 & 0.8306 & 0.0017 \\
29 & Spectfheart & 0.6835 & 0.0149 & 0.6408 & 0.0129 & 0.6431 & 0.0188 & 0.6015 & 0.0131 \\
30 & Tae & 0.5874 & 0.0125 & 0.5139 & 0.0285 & 0.5192 & 0.0319 & 0.5099 & 0.0207 \\
31 & Texture & 0.9889 & 0.0005 & 0.9845 & 0.0010 & 0.9814 & 0.0009 & 0.9766 & 0.0008 \\
32 & Thyroid & 0.9038 & 0.0012 & 0.8834 & 0.0020 & 0.8663 & 0.0016 & 0.8457 & 0.0019 \\
33 & Titanic & 0.7897 & 0.0009 & 0.7899 & 0.0013 & 0.7717 & 0.0049 & 0.7564 & 0.0010 \\
34 & Twonorm & 0.9381 & 0.0014 & 0.9194 & 0.0019 & 0.9006 & 0.0018 & 0.8880 & 0.0018 \\
35 & Vehicle & 0.6833 & 0.0060 & 0.6619 & 0.0102 & 0.6426 & 0.0078 & 0.6344 & 0.0093 \\
36 & Vowel & 0.9862 & 0.0027 & 0.9767 & 0.0024 & 0.9743 & 0.0023 & 0.9618 & 0.0028 \\
37 & Wdbc & 0.9499 & 0.0037 & 0.9387 & 0.0062 & 0.9250 & 0.0060 & 0.9178 & 0.0057 \\
38 & Wine & 0.9506 & 0.0069 & 0.9365 & 0.0046 & 0.9298 & 0.0130 & 0.9022 & 0.0113 \\
39 & Winequality-red & 0.6196 & 0.0052 & 0.5790 & 0.0080 & 0.5444 & 0.0032 & 0.5225 & 0.0061 \\
40 & Wisconsin & 0.9492 & 0.0022 & 0.9384 & 0.0031 & 0.9388 & 0.0041 & 0.9290 & 0.0053 \\ \hline
& Avg & 0.8027 & 0.0060 & 0.7835 & 0.0085 & 0.7677 & 0.0091 & 0.7547 & 0.0085 \\ \hline
\end{tabular}
\label{table7}
\end{table} 

\subsection{}
The detailed experimental results of IBT-U-K-S are given by Table 8.
\begin{table}\centering
\caption{The detailed experimental results of IBT-U-K-S.}
\begin{tabular}{llcccccccc}
\hline
\multirow{2}{*}{ID} & \multirow{2}{*}{Names} & \multicolumn{2}{c}{\emph{k}=3} & \multicolumn{2}{c}{\emph{k}=5}  & \multicolumn{2}{c}{\emph{k}=7} & \multicolumn{2}{c}{\emph{k}=9}  \\ \cline{3-10} 
              &        & \multicolumn{1}{c}{Avg} & \multicolumn{1}{c}{Std} & \multicolumn{1}{c}{Avg} & \multicolumn{1}{c}{Std} & \multicolumn{1}{c}{Avg} & \multicolumn{1}{c}{Std} & \multicolumn{1}{c}{Avg} & \multicolumn{1}{c}{Std} \\ \hline
1 & Appendicitis & 0.7585 & 0.0119 & 0.7594 & 0.0156 & 0.7896 & 0.0214 & 0.8047 & 0.0100 \\
2 & Balance & 0.7282 & 0.0047 & 0.7490 & 0.0070 & 0.7494 & 0.0069 & 0.7878 & 0.0083 \\
3 & Banana & 0.8826 & 0.0011 & 0.8885 & 0.0014 & 0.8941 & 0.0015 & 0.8965 & 0.0010 \\
4 & Bands & 0.6915 & 0.0097 & 0.6734 & 0.0129 & 0.6770 & 0.0100 & 0.6575 & 0.0129 \\
5 & Bupa & 0.6232 & 0.0189 & 0.6188 & 0.0140 & 0.6101 & 0.0101 & 0.6168 & 0.0118 \\
6 & Cleveland & 0.4879 & 0.0131 & 0.4845 & 0.0092 & 0.4916 & 0.0091 & 0.4889 & 0.0081 \\
7 & Dermatology & 0.9567 & 0.0020 & 0.9536 & 0.0024 & 0.9489 & 0.0042 & 0.9464 & 0.0018 \\
8 & Haberman & 0.6010 & 0.0104 & 0.6173 & 0.0117 & 0.6281 & 0.0111 & 0.6212 & 0.0147 \\
9 & Hayes-roth & 0.7325 & 0.0218 & 0.6038 & 0.0341 & 0.4988 & 0.0206 & 0.4850 & 0.0236 \\
10 & Heart & 0.7833 & 0.0066 & 0.7985 & 0.0063 & 0.8037 & 0.0086 & 0.8026 & 0.0065 \\
11 & Hepatitis & 0.7950 & 0.0087 & 0.8150 & 0.0053 & 0.8000 & 0.0118 & 0.8063 & 0.0106 \\
12 & Ionosphere & 0.8698 & 0.0030 & 0.8695 & 0.0042 & 0.8678 & 0.0047 & 0.8667 & 0.0032 \\
13 & Iris & 0.9587 & 0.0042 & 0.9600 & 0.0054 & 0.9593 & 0.0073 & 0.9587 & 0.0061 \\
14 & Led7digit & 0.7088 & 0.0049 & 0.7242 & 0.0075 & 0.7336 & 0.0075 & 0.7324 & 0.0065 \\
15 & Mammographic & 0.7760 & 0.0028 & 0.8037 & 0.0024 & 0.8060 & 0.0045 & 0.8083 & 0.0036 \\
16 & Marketing & 0.2922 & 0.0027 & 0.2996 & 0.0028 & 0.3052 & 0.0018 & 0.3084 & 0.0027 \\
17 & Monks-2 & 0.7752 & 0.0084 & 0.7426 & 0.0109 & 0.7384 & 0.0096 & 0.7153 & 0.0095 \\
18 & Movement\_libras & 0.7839 & 0.0083 & 0.7106 & 0.0095 & 0.6264 & 0.0079 & 0.5964 & 0.0113 \\
19 & Newthyroid & 0.9577 & 0.0056 & 0.9507 & 0.0050 & 0.9577 & 0.0056 & 0.9535 & 0.0066 \\
20 & Page-blocks & 0.8574 & 0.0012 & 0.8441 & 0.0013 & 0.8377 & 0.0016 & 0.8427 & 0.0010 \\
21 & Penbased & 0.9934 & 0.0002 & 0.9919 & 0.0003 & 0.9902 & 0.0002 & 0.9889 & 0.0003 \\
22 & Phoneme & 0.8736 & 0.0018 & 0.8656 & 0.0010 & 0.8568 & 0.0016 & 0.8506 & 0.0017 \\
23 & Pima & 0.7250 & 0.0045 & 0.7354 & 0.0069 & 0.7316 & 0.0052 & 0.7311 & 0.0067 \\
24 & Ring & 0.7155 & 0.0021 & 0.6885 & 0.0012 & 0.6687 & 0.0013 & 0.6539 & 0.0016 \\
25 & Satimage & 0.9024 & 0.0016 & 0.9021 & 0.0013 & 0.8988 & 0.0013 & 0.8959 & 0.0009 \\
26 & Segment & 0.9601 & 0.0014 & 0.9515 & 0.0015 & 0.9506 & 0.0018 & 0.9487 & 0.0016 \\
27 & Sonar & 0.8375 & 0.0101 & 0.8341 & 0.0129 & 0.7947 & 0.0072 & 0.7683 & 0.0136 \\
28 & Spambase & 0.9047 & 0.0018 & 0.9031 & 0.0011 & 0.9048 & 0.0013 & 0.9026 & 0.0017 \\
29 & Spectfheart & 0.6296 & 0.0101 & 0.5906 & 0.0092 & 0.5918 & 0.0092 & 0.5809 & 0.0076 \\
30 & Tae & 0.5318 & 0.0179 & 0.5252 & 0.0269 & 0.5152 & 0.0112 & 0.5099 & 0.0259 \\
31 & Texture & 0.9868 & 0.0004 & 0.9835 & 0.0007 & 0.9811 & 0.0005 & 0.9785 & 0.0007 \\
32 & Thyroid & 0.7707 & 0.0016 & 0.7826 & 0.0024 & 0.7568 & 0.0016 & 0.7504 & 0.0021 \\
33 & Titanic & 0.7601 & 0.0000 & 0.7607 & 0.0013 & 0.7883 & 0.0008 & 0.7892 & 0.0001 \\
34 & Twonorm & 0.9667 & 0.0008 & 0.9710 & 0.0005 & 0.9726 & 0.0005 & 0.9732 & 0.0007 \\
35 & Vehicle & 0.7116 & 0.0067 & 0.7047 & 0.0119 & 0.6974 & 0.0067 & 0.6918 & 0.0070 \\
36 & Vowel & 0.9606 & 0.0048 & 0.8629 & 0.0095 & 0.7551 & 0.0113 & 0.6969 & 0.0093 \\
37 & Wdbc & 0.9645 & 0.0026 & 0.9664 & 0.0021 & 0.9680 & 0.0031 & 0.9685 & 0.0029 \\
38 & Wine & 0.9534 & 0.0075 & 0.9528 & 0.0054 & 0.9517 & 0.0060 & 0.9573 & 0.0039 \\
39 & Winequality-red & 0.4826 & 0.0070 & 0.4994 & 0.0057 & 0.4946 & 0.0066 & 0.5063 & 0.0078 \\
40 & Wisconsin & 0.9750 & 0.0034 & 0.9755 & 0.0029 & 0.9739 & 0.0013 & 0.9735 & 0.0016 \\ \hline
& Avg & 0.7906 & 0.0059 & 0.7829 & 0.0068 & 0.7742 & 0.0061 & 0.7703 & 0.0064 \\ \hline
\end{tabular}
\label{table8}
\end{table}

\subsection{}

The detailed experimental results of TBC and IDC are given in Table 9. 
\begin{table}\centering
\caption{The detailed experimental results of TBC and IDC.}
\begin{tabular}{llcccc}
\hline
\multirow{2}{*}{ID} & \multirow{2}{*}{Names} & \multicolumn{2}{c}{IBC} & \multicolumn{2}{c}{IDC} \\ \cline{3-6} 
            &          & \multicolumn{1}{c}{Avg} & \multicolumn{1}{c}{Std} & \multicolumn{1}{c}{Avg} & \multicolumn{1}{c}{Std}  \\ \hline
1 & Appendicitis & 0.8613  & 0.0064  & 0.8075  & 0.0101 \\
2 & Balance & 0.8654  & 0.0050  & 0.7618  & 0.0065 \\
3 & Banana & 0.5568  & 0.0013  & 0.7313  & 0.0019 \\
4 & Bands & 0.6088  & 0.0115  & 0.5841  & 0.0141 \\
5 & Bupa & 0.6275  & 0.0088  & 0.5803  & 0.0086 \\
6 & Cleveland & 0 & 0 & 0.4892  & 0.0126 \\
7 & Dermatology & 0 & 0 & 0.8746  & 0.0066 \\
8 & Haberman & 0.7310  & 0.0064  & 0.6876  & 0.0222 \\
9 & Hayes-roth & 0.5288  & 0.0053  & 0.4744  & 0.0238 \\
10 & Heart & 0.8396  & 0.0072  & 0.8170  & 0.0040 \\
11 & Hepatitis & 0 & 0 & 0.8475  & 0.0211 \\
12 & Ionosphere & 0.8695  & 0.0057  & 0.7513  & 0.0043 \\
13 & Iris & 0.6667  & 0.0000  & 0.9060  & 0.0021 \\
14 & Led7digit & 0.2622  & 0.0109  & 0.4736  & 0.0080 \\
15 & Mammographic & 0.8088  & 0.0016  & 0.7982  & 0.0017 \\
16 & Marketing & 0.2652  & 0.0029  & 0.1284  & 0.0018 \\
17 & Monks-2 & 0.5294  & 0.0206  & 0.6391  & 0.0140 \\
18 & Movement\_libras & 0 & 0 & 0.2642  & 0.0169 \\
19 & Newthyroid & 0.3023  & 0.0000  & 0.8377  & 0.0056 \\
20 & Page-blocks & 0.0750  & 0.0006  & 0.8892  & 0.0007 \\
21 & Penbased & 0.1998  & 0.0000  & 0.6636  & 0.0004 \\
22 & Phoneme & 0.7595  & 0.0005  & 0.7684  & 0.0010 \\
23 & Pima & 0.7615  & 0.0041  & 0.7177  & 0.0028 \\
24 & Ring & 0.7621  & 0.0008  & 0.9603  & 0.0004 \\
25 & Satimage & 0.3448  & 0.0002  & 0.6317  & 0.0010 \\
26 & Segment & 0.2857  & 0.0000  & 0.6624  & 0.0022 \\
27 & Sonar & 0.7447  & 0.0159  & 0.7226  & 0.0116 \\
28 & Spambase & 0.9064  & 0.0011  & 0.8376  & 0.0006 \\
29 & Spectfheart & 0.6105  & 0.0122  & 0.7528  & 0.0138 \\
30 & Tae & 0.4974  & 0.0096  & 0.4020  & 0.0153 \\
31 & Texture & 0.3163  & 0.0013  & 0.5477  & 0.0022 \\
32 & Thyroid & 0.0713  & 0.0001  & 0.9239  & 0.0005 \\
33 & Titanic & 0.7807  & 0.0008  & 0.6900  & 0.0015 \\
34 & Twonorm & 0.9781  & 0.0003  & 0.9771  & 0.0001 \\
35 & Vehicle & 0.5324  & 0.0196  & 0.3018  & 0.0041 \\
36 & Vowel & 0.1818  & 0.0000  & 0.2899  & 0.0060 \\
37 & Wdbc & 0.9617  & 0.0023  & 0.9374  & 0.0017 \\
38 & Wine & 0.6011  & 0.0000  & 0.9472  & 0.0060 \\
39 & Winequality-red & 0 & 0 & 0.4021  & 0.0026 \\
40 & Wisconsin & 0.9581  & 0.0012  & 0.9555  & 0.0016 \\ \hline
& Avg & 0.5901  & 0.0047  & 0.6859  & 0.0066 \\ \hline
\end{tabular}
\label{table9}
\end{table}

\subsection{}

The detailed experimental results of \emph{k}-NN are given in Table 10. 
\begin{table}\centering
\caption{The detailed experimental results of \emph{k}-NN.}
\begin{tabular}{llcccccccc}
\hline
\multirow{2}{*}{ID} & \multirow{2}{*}{Names} & \multicolumn{2}{c}{\emph{k}=3} & \multicolumn{2}{c}{\emph{k}=5}  & \multicolumn{2}{c}{\emph{k}=7} & \multicolumn{2}{c}{\emph{k}=9}  \\ \cline{3-10} 
              &        & \multicolumn{1}{c}{Avg} & \multicolumn{1}{c}{Std} & \multicolumn{1}{c}{Avg} & \multicolumn{1}{c}{Std} & \multicolumn{1}{c}{Avg} & \multicolumn{1}{c}{Std} & \multicolumn{1}{c}{Avg} & \multicolumn{1}{c}{Std} \\ \hline
1 & Appendicitis & 0.8406  & 0.0094  & 0.8642  & 0.0119  & 0.8764  & 0.0030  & 0.8708  & 0.0100 \\
2 & Balance & 0.8485  & 0.0065  & 0.8661  & 0.0059  & 0.8813  & 0.0048  & 0.8928  & 0.0048 \\
3 & Banana & 0.8841  & 0.0014  & 0.8896  & 0.0012  & 0.8942  & 0.0021  & 0.8978  & 0.0012 \\
4 & Bands & 0.7093  & 0.0122  & 0.6942  & 0.0122  & 0.6797  & 0.0083  & 0.6712  & 0.0098 \\
5 & Bupa & 0.6371  & 0.0113  & 0.6078  & 0.0130  & 0.6238  & 0.0121  & 0.6293  & 0.0134 \\
6 & Cleveland & 0.5545  & 0.0152  & 0.5545  & 0.0057  & 0.5663  & 0.0115  & 0.5626  & 0.0117 \\
7 & Dermatology & 0.9623  & 0.0033  & 0.9592  & 0.0027  & 0.9575  & 0.0039  & 0.9517  & 0.0040 \\
8 & Haberman & 0.6954  & 0.0109  & 0.6944  & 0.0082  & 0.7111  & 0.0054  & 0.7186  & 0.0070 \\
9 & Hayes-roth & 0.6350  & 0.0187  & 0.5575  & 0.0255  & 0.4344  & 0.0215  & 0.3581  & 0.0228 \\
10 & Heart & 0.7778  & 0.0089  & 0.8033  & 0.0066  & 0.8126  & 0.0068  & 0.8115  & 0.0069 \\
11 & Hepatitis & 0.8288  & 0.0145  & 0.8525  & 0.0255  & 0.8800  & 0.0134  & 0.8563  & 0.0169 \\
12 & Ionosphere & 0.8570  & 0.0044  & 0.8501  & 0.0054  & 0.8393  & 0.0041  & 0.8425  & 0.0043 \\
13 & Iris & 0.9507  & 0.0034  & 0.9560  & 0.0034  & 0.9673  & 0.0066  & 0.9527  & 0.0049 \\
14 & Led7digit & 0.6598  & 0.0077  & 0.7116  & 0.0047  & 0.7090  & 0.0058  & 0.7234  & 0.0041 \\
15 & Mammographic & 0.7678  & 0.0055  & 0.7981  & 0.0067  & 0.7999  & 0.0051  & 0.8027  & 0.0050 \\
16 & Marketing & 0.2872  & 0.0030  & 0.2942  & 0.0015  & 0.2990  & 0.0025  & 0.3050  & 0.0020 \\
17 & Monks-2 & 0.7972  & 0.0072  & 0.8000  & 0.0054  & 0.7914  & 0.0127  & 0.7644  & 0.0074 \\
18 & Movement\_libras & 0.8075  & 0.0049  & 0.7417  & 0.0103  & 0.7181  & 0.0090  & 0.6739  & 0.0218 \\
19 & Newthyroid & 0.9409  & 0.0044  & 0.9381  & 0.0058  & 0.9316  & 0.0054  & 0.9237  & 0.0050 \\
20 & Page-blocks & 0.9596  & 0.0012  & 0.9583  & 0.0009  & 0.9545  & 0.0009  & 0.9536  & 0.0006 \\
21 & Penbased & 0.9935  & 0.0004  & 0.9926  & 0.0004  & 0.9919  & 0.0003  & 0.9905  & 0.0003 \\
22 & Phoneme & 0.8878  & 0.0021  & 0.8808  & 0.0028  & 0.8752  & 0.0017  & 0.8701  & 0.0023 \\
23 & Pima & 0.7396  & 0.0055  & 0.7367  & 0.0072  & 0.7449  & 0.0055  & 0.7357  & 0.0046 \\
24 & Ring & 0.7186  & 0.0014  & 0.6922  & 0.0010  & 0.6747  & 0.0012  & 0.6608  & 0.0017 \\
25 & Satimage & 0.9096  & 0.0012  & 0.9078  & 0.0011  & 0.9065  & 0.0015  & 0.9049  & 0.0019 \\
26 & Segment & 0.9613  & 0.0020  & 0.9532  & 0.0014  & 0.9502  & 0.0015  & 0.9481  & 0.0015 \\
27 & Sonar & 0.8303  & 0.0072  & 0.8135  & 0.0115  & 0.7880  & 0.0135  & 0.7457  & 0.0175 \\
28 & Spambase & 0.9019  & 0.0021  & 0.9030  & 0.0015  & 0.8995  & 0.0013  & 0.8959  & 0.0023 \\
29 & Spectfheart & 0.7150  & 0.0134  & 0.7390  & 0.0149  & 0.7629  & 0.0142  & 0.7547  & 0.0124 \\
30 & Tae & 0.5119  & 0.0153  & 0.5219  & 0.0184  & 0.5086  & 0.0253  & 0.4927  & 0.0263 \\
31 & Texture & 0.9878  & 0.0005  & 0.9853  & 0.0005  & 0.9828  & 0.0007  & 0.9809  & 0.0007 \\
32 & Thyroid & 0.9391  & 0.0008  & 0.9407  & 0.0005  & 0.9401  & 0.0005  & 0.9400  & 0.0002 \\
33 & Titanic & 0.6109  & 0.0107  & 0.7796  & 0.0118  & 0.7819  & 0.0013  & 0.7816  & 0.0034 \\
34 & Twonorm & 0.9650  & 0.0010  & 0.9697  & 0.0007  & 0.9705  & 0.0008  & 0.9714  & 0.0006 \\
35 & Vehicle & 0.7033  & 0.0051  & 0.7025  & 0.0054  & 0.7039  & 0.0055  & 0.6941  & 0.0096 \\
36 & Vowel & 0.9706  & 0.0025  & 0.9387  & 0.0057  & 0.8871  & 0.0071  & 0.7972  & 0.0108 \\
37 & Wdbc & 0.9692  & 0.0017  & 0.9678  & 0.0024  & 0.9705  & 0.0027  & 0.9692  & 0.0028 \\
38 & Wine & 0.9640  & 0.0039  & 0.9573  & 0.0089  & 0.9596  & 0.0052  & 0.9567  & 0.0088 \\
39 & Winequality-red & 0.5839  & 0.0062  & 0.5902  & 0.0069  & 0.5797  & 0.0040  & 0.5803  & 0.0042 \\
40 & Wisconsin & 0.9691  & 0.0022  & 0.9742  & 0.0024  & 0.9728  & 0.0019  & 0.9706  & 0.0021 \\ \hline
& Avg & 0.8058  & 0.0060  & 0.8085  & 0.0067  & 0.8045  & 0.0060  & 0.7951  & 0.0069 \\ \hline
\end{tabular}
\label{table10}
\end{table}

\subsection{}
The detailed experimental results of SVM and DT are given in Table 11. 
\begin{table}\centering
\caption{The detailed experimental results of SVM and DT.}
\begin{tabular}{llcccccc}
\hline
\multirow{2}{*}{ID} & \multirow{2}{*}{Names} & \multicolumn{2}{c}{SVM} & \multicolumn{2}{c}{DT} \\ \cline{3-6} 
&   & \multicolumn{1}{c}{Avg} & \multicolumn{1}{c}{Std} & \multicolumn{1}{c}{Avg} & \multicolumn{1}{c}{Std}  \\ \hline
1 & Appendicitis & 0.8736 & 0.0049 & 0.8358 & 0.0135 \\
2 & Balance & 0.8698 & 0.0060 & 0.7894 & 0.0080 \\
3 & Banana & 0.5517 & 0.0000 & 0.8799 & 0.0027 \\
4 & Bands & 0.6877 & 0.0107 & 0.6285 & 0.0272 \\
5 & Bupa & 0.5791 & 0.0018 & 0.6571 & 0.0183 \\
6 & Cleveland & 0.5859 & 0.0104 & 0.5091 & 0.0079 \\
7 & Dermatology & 0.9673 & 0.0019 & 0.9374 & 0.0058 \\
8 & Haberman & 0.7340 & 0.0017 & 0.6935 & 0.0139 \\
9 & Hayes-roth & 0.5144 & 0.0198 & 0.8181 & 0.0192 \\
10 & Heart & 0.8374 & 0.0041 & 0.7581 & 0.0196 \\
11 & Hepatitis & 0.8575 & 0.0278 & 0.8350 & 0.0269 \\
12 & Ionosphere & 0.8821 & 0.0054 & 0.8806 & 0.0101 \\
13 & Iris & 0.9613 & 0.0061 & 0.9487 & 0.0045 \\
14 & Led7digit & 0.7392 & 0.0075 & 0.7114 & 0.0075 \\
15 & Mammographic & 0.7959 & 0.0026 & 0.7988 & 0.0065 \\
16 & Marketing & 0.3210 & 0.0014 & 0.2970 & 0.0032 \\
17 & Monks-2 & 0.6713 & 0.0000 & 0.9067 & 0.0130 \\
18 & Movement\_libras & 0.7197 & 0.0117 & 0.6572 & 0.0265 \\
19 & Newthyroid & 0.8944 & 0.0062 & 0.9298 & 0.0060 \\
20 & Page-blocks & 0.9342 & 0.0005 & 0.9649 & 0.0010 \\
21 & Penbased & 0.9784 & 0.0004 & 0.9582 & 0.0010 \\
22 & Phoneme & 0.7731 & 0.0008 & 0.8650 & 0.0032 \\
23 & Pima & 0.7699 & 0.0032 & 0.7078 & 0.0105 \\
24 & Ring & 0.7651 & 0.0008 & 0.8858 & 0.0028 \\
25 & Satimage & 0.8646 & 0.0008 & 0.8608 & 0.0039 \\
26 & Segment & 0.9303 & 0.0012 & 0.9568 & 0.0039 \\
27 & Sonar & 0.7736 & 0.0169 & 0.7221 & 0.0185 \\
28 & Spambase & 0.9031 & 0.0009 & 0.9190 & 0.0028 \\
29 & Spectfheart & 0.7951 & 0.0018 & 0.7401 & 0.0155 \\
30 & Tae & 0.5364 & 0.0219 & 0.5444 & 0.0168 \\
31 & Texture & 0.9873 & 0.0003 & 0.9220 & 0.0030 \\
32 & Thyroid & 0.9371 & 0.0001 & 0.9960 & 0.0004 \\
33 & Titanic & 0.7760 & 0.0000 & 0.7898 & 0.0013 \\
34 & Twonorm & 0.9783 & 0.0003 & 0.8431 & 0.0048 \\
35 & Vehicle & 0.7356 & 0.0039 & 0.7139 & 0.0115 \\
36 & Vowel & 0.7129 & 0.0062 & 0.7666 & 0.0111 \\
37 & Wdbc & 0.9773 & 0.0027 & 0.9185 & 0.0040 \\
38 & Wine & 0.9860 & 0.0040 & 0.9096 & 0.0107 \\
39 & Winequality-red & 0.5841 & 0.0027 & 0.6077 & 0.0102 \\
40 & Wisconsin & 0.9687 & 0.0021 & 0.9492 & 0.0042 \\ \hline
& Avg & 0.7928  & 0.0050  & 0.8003  & 0.0095  \\ \hline
\end{tabular}
\label{table11}

\end{table}
\end{document}